\documentclass[conference]{IEEEtran}
\IEEEoverridecommandlockouts
\usepackage{cite}
\usepackage{amsmath,amssymb,amsfonts}
\usepackage{graphicx}
\usepackage{textcomp}
\usepackage{xcolor}
\usepackage{booktabs}
\usepackage{multirow}
\usepackage{array}
\PassOptionsToPackage{hyphens}{url}\usepackage{url}
\usepackage{balance}
\usepackage[colorlinks=true,linkcolor=blue,citecolor=blue,urlcolor=blue]{hyperref}
\usepackage{orcidlink}
\def\BibTeX{{\rm B\kern-.05em{\sc i\kern-.025em b}\kern-.08em
    T\kern-.1667em\lower.7ex\hbox{E}\kern-.125emX}}

\begin{document}

\title{Global AI Regulations for FAIR and Ethics in High-Risk Use Cases: A Comparative Review}

\author{
\IEEEauthorblockN{
  Aasish Kumar Sharma\,\orcidlink{0000-0002-7514-2340}\IEEEauthorrefmark{1},
  Dimitar Koysev\IEEEauthorrefmark{2},
  Christopher Anich\,\orcidlink{0009-0007-7423-3922}\IEEEauthorrefmark{3},
  Roshni Kumari Ojha\IEEEauthorrefmark{4},
  Julian Kunkel\,\orcidlink{0000-0002-6915-1179}\IEEEauthorrefmark{1}
}
\IEEEauthorblockA{
  \IEEEauthorrefmark{1}\textit{Mathematics and Computer Science, Georg-August-Universität Göttingen, Germany}
}
\IEEEauthorblockA{
  \IEEEauthorrefmark{2}\textit{Department of Economics, Burgas Free University, Bulgaria};
  kyosev.dimitar@gmail.com
}
\IEEEauthorblockA{
  \IEEEauthorrefmark{3}\textit{Medicine and Data Science, AnichLabs OÜ, Tallinn, Estonia};
  chris@anichlabs.com
}
\IEEEauthorblockA{
  \IEEEauthorrefmark{4}\textit{Faculty of Legal Studies, Sarala Birla University, India};
  sroshni3001@gmail.com
}
}

\maketitle

\begin{abstract}
AI governance is shifting from voluntary ethics to enforceable, risk-based regulation, yet cross-jurisdictional divergence creates compliance uncertainty for operators of high-stakes AI. We present a comparative matrix for the EU, US, and China that maps (i) risk classification triggers, (ii) binding obligations, (iii) enforcement and accountability mechanisms, and (iv) the degree to which FAIR principles are operationalised in practice. We stress-test the matrix on three high-impact domains: Electroencephalography (EEG)-guided rehabilitation robotics, AI-enabled debt collection in prospective Central Bank Digital Currency (CBDC) ecosystems, and AI-driven allocation of scarce Graphics Processing Unit (GPU) resources in emerging AI Factory infrastructures. Using primary legal texts and implementation evidence, we identify three recurring gaps: weak interoperability mandates, difficult operationalisation of cross-regime obligations (AI + sector regulation + data protection), and under-specified governance for critical digital infrastructure use cases. To bridge the implementation gap, we outline \emph{Knowledge Blocks}, a machine-checkable compliance artefact pattern based on Resource Description Framework/Web Ontology Language (RDF/OWL), Shapes Constraint Language (SHACL), and Provenance Ontology (PROV-O), enabling audit-ready compliance-by-design across multiple regimes.
\end{abstract}


\section{Introduction}

AI systems increasingly mediate access to essential services. When deployed in high-risk contexts, failures or bias may affect fundamental rights, safety, or societal trust, requiring governance that balances innovation with enforceable safeguards \cite{kusche2024possible}. The EU Artificial Intelligence Act (Regulation (EU)~2024/1689) introduces a risk-based horizontal framework \cite{act2025regulation}. The US relies on sector-specific legislation and voluntary guidance, reinforced by the January~2025 rescission of Executive Order~14110 \cite{house2025removing,blumenthal2024us}. China adopts a mandatory, state-centric model combining algorithmic registration and lifecycle oversight \cite{sheehan2023,filipova2024legal}. Despite international convergence on responsible AI \cite{unesco2021,oecd2019principles}, three recurring challenges persist: (i)~\emph{classification ambiguity} for systems such as AI-driven High-Performance Computing (HPC) resource allocation; (ii)~\emph{incomplete operationalisation} of FAIR principles (Findable, Accessible, Interoperable, Reusable) limiting auditability; and (iii)~\emph{enforcement asymmetry}, distorting incentives across regimes.

This paper addresses three research questions. \textbf{RQ1}: How do major AI governance regimes differ in classifying high-risk AI? \textbf{RQ2}: To what extent are FAIR principles embedded in binding obligations? \textbf{RQ3}: Which governance gaps persist across domains and jurisdictions? Our contributions are: \textbf{C1}~a reproducible jurisdiction-agnostic evaluation matrix; \textbf{C2}~cross-domain stress testing on three high-risk use cases exposing blind spots invisible in sector-isolated analyses; \textbf{C3}~identification and partial operationalisation of governance deficits through machine-checkable compliance artefacts (\emph{Knowledge Blocks}) \cite{Sharma2025EthicalAI}.

\section{Background and Related Work}

\subsection{High-Risk AI and Regulatory Philosophies}

AI regulation has shifted toward risk-based, enforceable frameworks. The EU AI Act defines high-risk systems via Annex~III, covering biometric identification, employment, credit, and critical infrastructure \cite{act2025regulation}. The US and China lack a unified definition but address elevated risk through sectoral and administrative mechanisms respectively \cite{davtyan2025us,sheehan2023china,filipova2024legal}. Governance philosophies diverge: the EU is rights-based and precautionary (ex ante); the US is market-driven and sectoral (ex post); China is state-centric and mandatory \cite{cole2024international}. The Organisation for Economic Co-operation and Development (OECD) and the United Nations Educational, Scientific and Cultural Organization (UNESCO) provide non-binding values-based frameworks \cite{oecd2019principles,unesco2021}, while the Council of Europe's Framework Convention on AI (CETS No.~225, 2024) signals a shift toward enforceability. Voluntary self-regulation is insufficient for high-risk AI \cite{ferretti2022institutionalist}.

\subsection{FAIR Principles and Machine-Checkable Compliance}

The FAIR principles (Findable, Accessible, Interoperable, Reusable) \cite{wilkinson2016fair} are increasingly relevant for AI governance as requirements for transparency and auditability. In high-risk AI, compliance depends on traceable data, accessible documentation, and portable compliance evidence. Even where documentation obligations exist, compliance artefacts are rarely machine-checkable, weakening cross-border auditability, especially for Interoperability \cite{ferretti2022institutionalist}. \emph{Knowledge Blocks} address this: modular compliance artefacts using RDF/OWL for representation, SHACL for validation, and PROV-O for provenance, enabling automated conformity checks and evidence graphs \cite{Sharma2025EthicalAI}. Domain-specific challenges include multi-layered obligations in healthcare AI (safety + data protection + AI Act) \cite{bessler2021safety,fu2022myoelectric}, fairness and due-process risks in financial AI and CBDCs \cite{krause2024future,cao2023fintech}, and the absence of AI-specific governance for HPC resource allocation despite high-stakes fairness effects.

\section{Methodology}

We adopt a structured comparative design examining three high-risk use cases across the EU, US, and China, with the EU AI Act as a consistent classification reference. Use cases were selected for (i)~clear high-risk characteristics under the EU AI Act and (ii)~functional relevance across jurisdictions: \textbf{UC1}~EEG-guided rehabilitation robotics (healthcare); \textbf{UC2}~AI-enabled debt collection in CBDC systems (finance); \textbf{UC3}~AI-driven resource allocation in AI Factories (HPC infrastructure) \cite{euaiact_timeline_ec}. The unit of analysis is the \emph{binding regulatory obligation}. For each use case and jurisdiction, obligations are coded along four dimensions: (1)~risk classification trigger, (2)~mandatory lifecycle controls, (3)~enforcement and accountability mechanisms, and (4)~operational FAIR support. Comparative matrices are constructed for each use case to identify classification ambiguity, implementation gaps, and enforcement asymmetries. Analysis is grounded in primary legal instruments complemented by secondary legal and technical analyses.

\section{Use Case Analysis}

\subsection{UC1: EEG-Guided Robotic Neurorehabilitation (Healthcare)}

UC1 considers a robotic arm for neurorehabilitation combining a wearable exoskeleton with EEG to infer motor intent. The Machine Learning (ML) component maps EEG patterns to actuator commands within a human-in-the-loop workflow. EEG data constitutes health and biometric information, so governance turns on medical device classification and special-category data processing \cite{world2014global,regulation2016regulation}.

\textit{EU:} UC1 qualifies as high-risk under Article~6(1) of the AI Act, triggering mandatory requirements on risk management, data governance, technical documentation, transparency, human oversight, and robustness \cite{act2025regulation}. These duties apply concurrently with the Medical Device Regulation (MDR) \cite{eumdr2017} and the General Data Protection Regulation (GDPR) \cite{regulation2016regulation}. Relevant standards include IEC~60601, IEC~80601-2-78, IEC~62304, ISO~14971, and ISO/IEC~42001 (see \url{https://www.iso.org} and \url{https://webstore.iec.ch}).

\textit{US:} Governed through the Food and Drug Administration (FDA) medical device pathway; AI-specific guidance is largely non-binding. The Health Insurance Portability and Accountability Act (HIPAA) addresses health data in covered-entity contexts. Fairness and lifecycle controls rely on voluntary frameworks.

\textit{China:} Combines National Medical Products Administration (NMPA) medical device regulation with algorithmic governance. The Personal Information Protection Law (PIPL) \cite{pipl2021} treats EEG data as sensitive personal information with strict processing and cross-border transfer constraints.

A limitation across all jurisdictions is the absence of machine-checkable compliance artefacts for documentation, data provenance, and human oversight verification. Table~\ref{tab:healthcare} summarizes the comparison.

\begin{table*}[t]
\centering
\caption{Regulatory Comparison: EEG-Guided Robotic Neurorehabilitation (UC1)}
\label{tab:healthcare}
\scriptsize
\begin{tabular}{p{11em}p{18em}p{18em}p{18em}}
\toprule
\textbf{Dimension} & \textbf{EU} & \textbf{US} & \textbf{China} \\
\midrule
\textbf{Legal Status} & Mandatory (AI Act + MDR + GDPR) & Mandatory device regulations; voluntary AI guidance & Mandatory (NMPA; CAC obligations conditional) \\
\midrule
\textbf{High-Risk Classification} & Explicit (Art.~6(1): AI as safety component) & Implicit (Class II/III medical device) & Device classification; algorithm review where applicable \\
\midrule
\textbf{Fairness Requirements} & Data governance and bias risk mitigation mandatory & Voluntary guidelines only & Explicit anti-discrimination in design, data, and models \\
\midrule
\textbf{Transparency} & Instructions for use, explanations required & Labeling requirements; voluntary AI disclosure & Algorithmic disclosure to authorities where required \\
\midrule
\textbf{Human Oversight} & Mandatory (Art.~14) & Clinical oversight required by practice standard & Required in clinical settings \\
\midrule
\textbf{Data Protection} & GDPR (Art.~9 special-category health data) & HIPAA (covered entities) & PIPL (strict health data controls) \\
\midrule
\textbf{Pre-Market Assessment} & Notified body + conformity assessment & FDA 510(k) or PMA & NMPA approval; additional algorithm review where applicable \\
\midrule
\textbf{Post-Market Monitoring} & Mandatory (MDR surveillance; AI Act monitoring obligations) & FDA adverse event reporting & NMPA vigilance + ongoing algorithm monitoring \\
\midrule
\textbf{Standards Referenced} & IEC 60601, 80601-2-78, 62304, ISO 14971, 42001 & IEC 60601, FDA guidance & GB standards (Chinese equivalents) \\
\midrule
\textbf{Penalties} & Administrative fines; MDR sanctions & Warning letter, recall, criminal liability (fraud) & Fines, license revocation, personal liability \\
\midrule
\textbf{Innovation Impact} & High compliance burden; harmonized EU market & Moderate burden; FDA pathway established & High burden; layered approvals \\
\midrule
\textbf{FAIR Support} & Strong (documentation, data governance) & Moderate (device labeling) & Strong where algorithm governance applies \\
\bottomrule
\end{tabular}
\end{table*}

\subsection{UC2: AI-Enabled Debt Collection in CBDC Systems (Finance)}

UC2 examines AI-enabled debt collection within a prospective CBDC ecosystem, where programmable money and real-time settlement support automated enforcement. High-risk characteristics arise when AI influences creditworthiness, payment constraints, or enforcement priorities, amplifying discrimination risks and raising due-process concerns \cite{krause2024future,allen2020design,cao2023fintech}.

\textit{EU:} UC2 intersects the EU AI Act, the Digital Euro legislative package \cite{digitaleurobill}, GDPR, and payment regulation \cite{regulation2016regulation,allen2020design}. AI-enabled debt collection plausibly triggers high-risk classification where it affects access to essential services. The key compliance challenge is \emph{regime coupling}: obligations from AI law, payment services law, and data protection must be operationalised coherently at runtime.

\textit{US:} No deployed CBDC; governance is mediated through sectoral statutes (Fair Debt Collection Practices Act (FDCPA), Fair Credit Reporting Act (FCRA), Equal Credit Opportunity Act (ECOA)) and Federal Trade Commission (FTC)/Consumer Financial Protection Bureau (CFPB) enforcement \cite{greenwald2022digital,xie2025divergent}. These provide ex post accountability but limited AI-specific ex ante lifecycle obligations.

\textit{China:} The e-CNY (Electronic Chinese Yuan, China's deployed CBDC) provides the most mature CBDC reference, with strong administrative oversight \cite{cheng2023decoding,liu2025impact,huang2022digital}. Where UC2 relies on algorithmic profiling, Cyberspace Administration of China (CAC) filing requirements may apply, supplemented by PIPL and cybersecurity controls \cite{cyberlaw2026}.

The key unresolved issue across jurisdictions is how to impose effective constraints on programmable enforcement while preserving contestability, proportionality, and non-discrimination. Table~\ref{tab:cbdc} summarizes UC2.

\begin{table*}[t]
\centering
\caption{Regulatory Comparison: CBDC and AI-Enabled Debt Collection (UC2)}
\label{tab:cbdc}
\scriptsize
\begin{tabular}{p{11em}p{18em}p{18em}p{18em}}
\toprule
\textbf{Dimension} & \textbf{EU} & \textbf{US} & \textbf{China} \\
\midrule
\textbf{CBDC Status} & Proposed (Digital Euro Package) & Exploratory only & Deployed (e-CNY) \\
\midrule
\textbf{AI Regulation} & AI Act (if high-risk) + Digital Euro Regulation & Sectoral laws (FDCPA, ECOA, FCRA) & CAC filing + PIPL + Cybersecurity Law \\
\midrule
\textbf{High-Risk Classification} & Likely (critical infrastructure, access to services) & No unified classification & Explicit (algorithm filing where applicable) \\
\midrule
\textbf{Fairness Requirements} & Data governance, bias detection (if high-risk) & Anti-discrimination laws (reactive) & Mandatory prevention across lifecycle \\
\midrule
\textbf{Transparency} & Documentation, explanations required & FCRA disclosure requirements & Mandatory algorithmic disclosure \\
\midrule
\textbf{Data Protection} & GDPR + PSD2 (Payment Services Directive 2) consumer protection & Limited federal privacy law & PIPL (comprehensive, strict) \\
\midrule
\textbf{Programmability} & Cautious; consumer protection paramount & Not applicable (no CBDC) & Extensive; state-controlled \\
\midrule
\textbf{User Rights} & GDPR rights; appeals required if high-risk & FDCPA dispute rights & View tags, modify preferences, challenge decisions \\
\midrule
\textbf{Enforcement} & Administrative supervision (AI + payment regulation) & FTC/CFPB enforcement & Administrative enforcement; platform sanctions \\
\midrule
\textbf{Regulatory Philosophy} & Rights-based, cautious innovation & Market-driven, stability-focused & State-centric, deployment-oriented \\
\midrule
\textbf{FAIR Support} & Strong (if high-risk); documentation mandates & Minimal (voluntary disclosure) & Strong (registries where applied) \\
\bottomrule
\end{tabular}
\end{table*}

\subsection{UC3: AI-Driven Resource Allocation in AI Factories (HPC)}

UC3 concerns AI-assisted allocation of scarce GPU resources in AI Factory infrastructures. ML-based allocation introduces opacity and feedback loops that may reinforce institutional hierarchies or disadvantage emerging research groups. Governance extends beyond efficiency to fairness, contestability, and accountability \cite{eurohpc2025,bhuyan2025european,top500_2025_11}.

\textit{EU:} Annex~III of the AI Act covers critical infrastructure and access to essential services but does not explicitly include HPC allocation \cite{act2024annex}. High-risk classification is plausible via foreseeable misuse and systemic impact on access or career outcomes \cite{act2025regulation}. Under such classification, obligations would include data governance, bias monitoring, transparency, human oversight, and contestation rights.

\textit{US:} No horizontal AI regime for HPC scheduling. Governance is primarily institutional, supported by internal policies and voluntary frameworks such as the National Institute of Standards and Technology (NIST) AI Risk Management Framework (AI RMF) \cite{nist2023airmf}. Fairness depends on local policy and audit capacity.

\textit{China:} Algorithmic governance may apply where systems have public or societal impact, requiring registration and anti-discrimination measures \cite{chinalawtranslate2022provisions,sheehan2022china}. Application to HPC allocation remains indirect, and bias attribution is technically difficult due to dependence on workload characteristics and institutional history.

The main bottleneck is not the absence of fairness principles but the lack of standardized, machine-checkable compliance artefacts for allocation decisions. Knowledge Blocks can encode inputs, constraints, explanations, and appeal traces as auditable semantic artefacts \cite{Sharma2025EthicalAI}. Table~\ref{tab:hpc} summarizes UC3.

\begin{table*}[t]
\centering
\caption{Regulatory Comparison: AI-Driven HPC Resource Allocation (UC3)}
\label{tab:hpc}
\scriptsize
\begin{tabular}{p{11em}p{18em}p{18em}p{18em}}
\toprule
\textbf{Dimension} & \textbf{EU} & \textbf{US} & \textbf{China} \\
\midrule
\textbf{Infrastructure} & 19 AI Factories, €10B investment & DOE/NSF facilities, NAIRR pilot & National-level HPC centers \\
\midrule
\textbf{Regulatory Status} & AI Act in force; UC3 scope unclear & No horizontal AI regime & Algorithmic governance provisions active \\
\midrule
\textbf{High-Risk Classification} & Potentially (Annex~III); plausible via misuse/impact & No unified classification & Case-dependent (registration triggers) \\
\midrule
\textbf{Fairness Requirements} & If high-risk: data governance and bias monitoring & Voluntary guidance; local policy & Anti-discrimination principles; operationalization hard \\
\midrule
\textbf{Transparency} & If high-risk: documentation and explanations & Internal policies only & Registration/disclosure where applicable \\
\midrule
\textbf{Algorithm Registration} & No general registry & No requirement & CAC filing for covered algorithms \\
\midrule
\textbf{User Rights} & If high-risk: information and contestation hooks & Internal complaint procedures & Rights depend on applicability of algorithm rules \\
\midrule
\textbf{Bias Monitoring} & If high-risk: required & Voluntary & Required where applicable; attribution challenging \\
\midrule
\textbf{FAIR Alignment} & Partial; interoperability under-specified & No mandate & Stronger on registrability; interoperability limited \\
\bottomrule
\end{tabular}
\end{table*}

\section{Cross-Cutting Analysis}

\subsection{Regulatory Philosophy Comparison}

Table~\ref{tab:philosophy} summarizes the dominant governance logics, which determine what becomes enforceable and how readily evidence can be produced across borders.

\begin{table}[ht]
\centering
\caption{Regulatory Philosophy Comparison}
\label{tab:philosophy}
\scriptsize
\begin{tabular}{p{4.5em}p{8em}p{8.5em}p{8em}}
\toprule
\textbf{Aspect} & \textbf{EU} & \textbf{US} & \textbf{China} \\
\midrule
\textbf{Foundation} & Rights-based, precautionary & Market-driven, innovation & State-centric, control \\
\midrule
\textbf{Approach} & Ex ante, horizontal & Ex post, sectoral & Comprehensive, mandatory \\
\midrule
\textbf{Enforcement} & Strong & Moderate (agencies) & Strong \\
\midrule
\textbf{Priority} & Safety + rights & Innovation + stability & Security + social values \\
\bottomrule
\end{tabular}
\end{table}

\subsection{FAIR Support Across Jurisdictions}
Across UC1 to UC3, FAIR support is uneven. Findability is strongest in China (registries), conditional in the EU, and largely voluntary in the US. Accessibility is explicit in China, conditional in the EU, and fragmented in the US. Interoperability remains the weakest dimension: compliance evidence is rarely portable or machine-checkable across any jurisdiction. Reusability is comparatively stronger in the EU and China through documentation mandates, but lacks standardized formats.

\subsection{Cross-Cutting Gaps}

\textit{G1: Classification and Scope Uncertainty (UC2, UC3):} UC3 (HPC allocation) is a boundary case where systemic impact supports a high-risk interpretation not explicitly covered by Annex~III. UC2 reflects a plausible high-risk trajectory still pending legislative resolution.

\textit{G2: Operationalisation Deficit (all UCs):} Obligations specify requirements but provide limited implementation guidance. Fairness metrics for UC3 are not standardized, explainability for EEG inference in UC1 remains underspecified, and UC2 lacks concrete constraints for automated enforcement.

\textit{G3: Enforcement Heterogeneity (all UCs):} The EU emphasizes ex ante compliance, China combines registration with administrative control, and the US relies on ex post sectoral enforcement, creating incompatible evidence and audit expectations in multinational settings.

\textit{G4: Missing Machine-Checkable Compliance Artefacts (FAIR-I):} Compliance documentation is rarely structured for automated validation or cross-border portability, motivating compliance-by-design approaches such as Knowledge Blocks \cite{Sharma2025EthicalAI}.

\section{Discussion}

\subsection{Insights from the Use Cases}
The three use cases expose distinct governance patterns. UC1 shows that AI in regulated products benefits from clear classification and established conformity pathways, though multi-layer compliance (AI Act + MDR + GDPR) creates integration burdens with no single auditing standard. UC2 highlights regime coupling in financial AI: when debt collection logic is embedded in a CBDC, obligations from AI regulation, payment services law, and data protection must be operationalised coherently at runtime, yet no current framework provides joint compliance guidance. UC3 reveals a governance gap for AI-driven infrastructure allocation: despite high-stakes effects on research access and career outcomes, classification is ambiguous (EU), internalized (US), or indirect (China). Across all three cases, ethical gaps are most visible where AI mediates access to services or infrastructure, and the central limitation is not a lack of declared principles but the absence of auditable, implementable enforcement mechanisms.

\subsection{Research Question Synthesis}

\textit{RQ1 (How do regimes differ in classifying high-risk AI?):} The EU follows preventive lifecycle governance through a horizontal risk-based framework; China emphasizes mandatory administrative registration and oversight; the US relies on sectoral, reactive enforcement. These differences shape both what is classified as high-risk and when obligations apply. Infrastructure AI (UC3) exposes the limits of all three frameworks: Annex~III is under-specified, US governance is institutional, and Chinese provisions apply indirectly.

\textit{RQ2 (To what extent are FAIR principles embedded in binding obligations?):} FAIR support is asymmetric. Findability is strongest in China through algorithm registries, conditional in the EU, and largely voluntary in the US. Accessibility is explicit in China, conditional in the EU (if high-risk), and fragmented in the US across sectoral enforcement. Reusability is comparatively stronger where documentation mandates exist (EU, China). Interoperability remains the weakest dimension across all jurisdictions: compliance artefacts are rarely standardized, machine-readable, or portable.

\textit{RQ3 (Which governance gaps and enforcement asymmetries persist?):} Four gaps recur: (G1) classification ambiguity for AI in critical infrastructure and novel financial systems; (G2) operationalisation deficits where obligations specify goals but not implementation; (G3) enforcement heterogeneity creating incompatible compliance expectations in multinational deployments; and (G4) the systemic absence of machine-checkable compliance artefacts limiting cross-border auditability.

\subsection{Limitations}

The study is qualitative and constrained by rapidly evolving regulation and the gap between formal rules and enforcement practice. It covers three use cases; broader generalization requires empirical validation, particularly for fairness measurement in infrastructure systems. Enforcement data, especially for China, relies on secondary sources.

\section{Conclusion and Recommendations}

AI governance remains fragmented despite convergence in principles. Classification ambiguity for emerging use cases, uneven FAIR operationalisation, and incompatible enforcement models limit the effectiveness of current regimes. The absence of machine-interpretable compliance mechanisms is the central technical gap, and it is particularly pressing ahead of EU AI Act full applicability in August~2026 \cite{euaiact_timeline_ec} for large-scale infrastructures such as AI Factories.

\textit{Recommendations:} Operators should perform early classification assessments, ensure traceable data-model-decision links, and implement continuous bias monitoring. Policymakers should clarify high-risk classification for infrastructure AI and mandate interoperable, machine-checkable compliance artefacts as audit standards. Research should prioritize standardized fairness metrics and semantic auditing tools for infrastructure-level AI. Future work will deploy Knowledge Blocks in operational settings, define semantic schemas for portable compliance evidence, and evaluate their role in cross-border compliance interoperability.

\section*{Acknowledgments}
{\footnotesize\textit{\copyright~2026~IEEE. Personal use of this material is permitted. Permission from IEEE must be obtained for all other uses. Accepted at the 50th IEEE Computers, Software, and Applications Conference (COMPSAC 2026), Madrid, Spain, July 7--10, 2026.}}

\renewcommand{\footnotesize}{\tiny}
This work was supported by NHR at GWDG and conducted within the KISSKI project (BMBF grant no.\ 01\,IS\,22\,093\,A-E) at the University of Göttingen. AI Factory governance activities are carried out within KISSKI and HammerHAI, Germany's EuroHPC AI Factory at HLRS, where GWDG is a consortium partner.\footnote{NHR: Nationales Hochleistungsrechnen; GWDG: Gesellschaft für wissenschaftliche Datenverarbeitung mbH Göttingen; KISSKI: KI-Servicezentrum für sensible und kritische Infrastruktur; BMBF: Bundesministerium für Bildung und Forschung; EuroHPC: European High Performance Computing Joint Undertaking; HLRS: High-Performance Computing Center Stuttgart.}
\renewcommand{\footnotesize}{\small}

\bibliographystyle{ieeetr}
\bibliography{references}

\balance
\renewcommand{\footnotesize}{\tiny}
\appendix

\section{EU AI Act High-Risk Classification Criteria}

Article~6 of Regulation (EU)~2024/1689 \cite{act2025regulation} defines high-risk AI via two pathways.

\textit{Pathway~1 (Art.~6(1)):}
AI systems used as safety components in products regulated under Annex~I\footnote{Annex~I of the EU AI Act lists regulated product categories including medical devices, machinery, toys, aviation, automotive, and radio equipment. Full text: \url{https://artificialintelligenceact.eu/annex/1/}} of the AI Act (e.g., medical devices, machinery, aviation, automotive) that require third-party conformity assessment under applicable Union harmonisation legislation.

\textit{Pathway~2 (Art.~6(2)):}
AI systems listed in Annex~III \cite{act2024annex} across eight domains: biometric identification; critical infrastructure; education; employment; access to essential services; law enforcement; migration and border control; and administration of justice and democratic processes. Annex~III systems may be exempt if limited to preparatory or supportive roles without influencing outcomes, except where profiling of natural persons is involved \cite{act2025regulation}.

\section{Chinese Algorithm Filing Statistics}

According to the Cyberspace Administration of China (CAC) algorithm registry\footnote{CAC Algorithm Registry: \url{https://www.cac.gov.cn/}. Filing requirements are defined in the Provisions on the Management of Algorithmic Recommendations (2022) \cite{chinalawtranslate2022provisions} and subsequent deep synthesis and generative AI provisions.} (January~2026 \cite{sheehan2022china}), over 1{,}400 algorithms from 450+ entities have been filed. Categories include recommendation (64\%), search/filtering (18\%), generative AI (12\%), and others (6\%). Major registrants include Tencent, Alibaba, ByteDance, Baidu, and JD.com \cite{sheehan2022china}. Filings require algorithm classification, application domain, core logic, training data sources, and a self-assessment report \cite{chinalawtranslate2022provisions}. Enforcement actions (2024 to 2025) include platform suspensions and administrative fines ranging from Renminbi (RMB)~100{,}000 to RMB~15~million under the Cybersecurity Law \cite{cyberlaw2026}.

\end{document}